\documentclass[applsci,article,accept,pdftex,moreauthors]{Definitions/mdpi} 
\usepackage{xcolor}
\firstpage{1} 
\pubvolume{1}
\issuenum{1}
\articlenumber{0}
\pubyear{2024}
\copyrightyear{2024}
\externaleditor{Academic Editor: Andrea Prati}
\datereceived{7 May 2024 } 
\daterevised{1 June 2024 } 
\dateaccepted{2 June 2024} 
\datepublished{18 June 2024} 
\hreflink{https://doi.org/} 

\Title{{Haze-Aware} Attention Network for Single-Image Dehazing} 

\Title{Haze-Aware Attention Network for Single-Image Dehazing}

\Author{Lihan Tong 
 $^{1}$, Yun Liu $^{2}$\orcidA{}, Weijia Li $^{3}$\orcidB{}, Liyuan Chen $^{1}$ and Erkang Chen $^{1,*}$\orcidC{}}

\AuthorNames{Lihan Tong, Yun Liu, Weijia Li, Liyuan Chen and Erkang Chen}

\AuthorCitation{Tong, 
 L.; Liu, Y.; Li, W.; Chen, L.; Chen, E.}

\address{%
$^{1}$ \quad School of Ocean Information Engineering, Jimei University, Xiamen 361021, China; \linebreak 202121301035@jmu.edu.cn (L.T.); lyuanchen@jmu.edu.cn (L.C.)\\
$^{2}$ \quad College of Artificial Intelligence, Southwest University, Chongqing 400715, China; yunliu@swu.edu.cn \\
$^{3}$ \quad School of Computer Science, Jimei University, Xiamen 361021, China; vjiali@jmu.edu.cn
}

\corres{Correspondence: ekchen@jmu.edu.cn}

\abstract{Single-image dehazing is a pivotal challenge in computer vision that seeks to remove haze from images and restore clean background details. Recognizing the limitations of traditional physical model-based methods and the inefficiencies of current attention-based solutions, we propose a new dehazing network combining an innovative Haze-Aware Attention Module (HAAM) with a Multiscale Frequency Enhancement Module (MFEM). The HAAM is inspired by the atmospheric scattering model, thus skillfully integrating physical principles into high-dimensional features for targeted dehazing. It picks up on latent features during the image restoration process, which gives a significant boost to the metrics, while the MFEM efficiently enhances high-frequency details, thus sidestepping wavelet or Fourier transform complexities.  It employs multiscale fields to extract and emphasize key frequency components with minimal parameter overhead. Integrated into a simple U-Net framework, our Haze-Aware Attention Network (HAA-Net) for single-image dehazing significantly outperforms existing attention-based and transformer models in efficiency and effectiveness. Tested across various public datasets, the HAA-Net sets new performance benchmarks. Our work not only advances the field of image dehazing but also offers insights into the design of attention mechanisms for broader applications in computer vision.}

\keyword{single-image \textls[-15]{dehazing; haze-aware attention; multiscale frequency enhancement module}} 
\begin{document}


\section{Introduction}
Single-image dehazing~\cite{aod,ffa-net,wu2021contrastive} aims to eliminate haze from images, thus accurately restoring the details of a clean background. This process is recognized as a classic example of an ill-posed problem, where the solution is not unique. Despite these challenges, single-image dehazing is a key area of research due to its wide range of applications in many computer vision tasks. In numerous computer vision tasks, these include  outdoor surveillance~\cite{ye2021perceiving}, outdoor scene understanding~\cite{sakaridis2018model,sakaridis2018semantic},  and object detection~\cite{li2018end,chen2018domain}.
The visual effect of fog is crucial for improving the accuracy and effectiveness of these tasks. Therefore, the pursuit of effective single-image dehazing methods has become a focal point of research. Over the past decade, this field has attracted a lot of interest from researchers and engineers, thus leading to the development of various innovative technologies and algorithms. These efforts are driven by the potential benefits that dehazing can bring to many applications, thus making it a dynamic and vibrant area of study within the broader field of computer~vision.

In recent years, the swift advancement in deep learning technologies has brought attention-based dehazing networks to the forefront of interest. The key to their growing popularity lies in the attention module's ability to selectively target various areas and channels. This adaptability is especially valuable for dehazing networks, given the uneven spatial spread of haze degradation. 
 Such attention modules offer a tailored approach to effectively restoring clean images, thus addressing the specific challenges posed by haze. The attention-based dehazing methods~\cite{ffa-net} have achieved performance far beyond that of traditional physical model-based methods~\cite{aod}. The PCFA module~\cite{zhang2020pyramid}  taps feature pyramid and channel attention to extract and focus on crucial image features for effective dehazing.  Zhang et al.~\cite{zhang2020multi} introduced an RMAM with an attention block to help networks concentrate on key features during learning. Zhang et al.~\cite{9381290} designed a network that teams up multilevel feature blending with mixed convolution attention to steadily and smartly boost dehazing results. However, the explanation behind these attention-based approaches remains unclear, as they do not have a solid link to the atmospheric scattering model ~\cite{dcp,aod}. Additionally, some of these techniques ~\cite{ffa-net,song2023vision,griddehazenet} employ intricate designs or self-attention mechanisms, thus leading to suboptimal efficiency.  In this study, we take a fresh look at the attention mechanism designs in dehazing networks and introduce a new approach inspired by physical priors~\cite{mccartney1976optics,narasimhan2000chromatic,narasimhan2002vision}, which is named the Haze-Aware Attention Module (HAAM). This module ingeniously applies the attention mechanism to mimic the parameters of the atmospheric scattering model, thus expressing these parameters through high-dimensional features. By employing these features constrained  by the physical model, we conducted the dehazing process within the feature space, thus achieving outstanding results in both performance \mbox{and efficiency}.  

In addition to HAAM, we also  developed a new Multiscale Frequency Enhancement Module (MFEM) designed to boost high-frequency details without relying on wavelet~\cite{zou2021sdwnet} or Fourier transforms~\cite{mao2021deep}. MFEM uses a 4-scale receptive field to extract contextual features, thus fully adapting to receptive fields of different sizes. It also emphasizes important information in the frequency domain through lightweight kernel learning parameters in the channel dimension, thus effectively enhancing the dehazing effect. This approach sidesteps the extra computational work typically needed for these methods' reverse transformations, thus resulting in an efficient and stable enhancement of features. By combining our proposed HAAM and MFEM with the straightforward U-Net architecture, we have created the Haze-Aware Attention Network (HAA-Net) for single-image dehazing. Our method has been tested on both synthetic and real-world datasets. In terms of metrics and visual effects, our approach significantly outperformed traditional attention-based methods, as well as transformer-based methods.

The contributions of this work are summarized as follows:
\begin{itemize}
    \item We developed an efficient attention mechanism known as the HAAM, which is inspired by the atmospheric scattering model that smartly incorporates physical principles into high-dimensional features.

 \item We crafted a multiscale frequency enhancement module that tunes high-frequency features, thus effectively bringing back the finer details of hazy images.

 \item Our HAA-Net set new benchmarks in performance across several public datasets. Notably, it reached the PSNR/SSIM of 41.23 dB/0.996 on the RESIDE-Indoor dataset, thus showcasing its exceptional dehazing performance. 
\end{itemize}

\section{Related Work}
\subsection{Prior-Based Image Dehazing}
Research on single-image dehazing in computer vision and computer graphics has been widely explored. Traditional methods have relied on  priors such as the dark channel prior (DCP)~\cite{dcp}, color attenuation prior ~\cite{zhu2014single}, and nonlocal prior~\cite{berman2016non} to estimate scattering light, atmospheric light, depth, and transmission map ~\cite{mccartney1976optics,tan2008visibility,dcp,zhu2014single,berman2016non}. These methods are backed by strong principles and are interpretable. However, they may not perform well in real-world image dehazing scenarios because they only extract features based on the atmospheric scattering model at the image level, without accessing deep latent features.
The Atmospheric Scattering Model (ASM) has been the cornerstone for many previous works in constructing image dehazing networks. These methods have explicitly incorporated the ASM to enhance the generalization capability of their models, thus thoroughly validating the effectiveness of the ASM. In contrast to these approaches, we introduced the ASM at the feature level, thus leveraging it to learn more latent features. Additionally, we allocated a substantial number of channels for the atmospheric light value A, thus aiming to better adapt to the complexities of real-world scenarios.

\subsection{Deep Learning-Based Image Dehazing}
Due to the inability of prior-based dehazing methods to adapt well to all haze scenes, recent dehazing efforts have moved away from using priors. 
Some end-to-end networks directly estimated haze-free images ~\cite{ren2018gated,griddehazenet,qu2019enhanced,ffa-net,msbdn,wu2021contrastive,9932546,cui2022selective}. AECRNet, SFNet, and others use a U-shaped structure, which has been proven to be superior for haze removal. These methods have achieved some results, but they perform poorly in dehazing real images. Recently, transformers 
~\cite{liu2021swin,zamir2022restormer,tsai2022stripformer,zhu2023biformer,liu2023nighthazeformer} have been used in image tasks due to their advantage in capturing long-range relationships. However, their computational complexity increases quadratically with resolution, thereby making them unsuitable for pixel-to-pixel tasks like dehazing. Moreover, these methods lack theoretical interpretability. Therefore, instead of using transformers, we developed our own more efficient attention mechanism based on physical~priors.

\subsection{Attention-Based Image Dehazing}
Attention mechanisms have been playing a crucial role in the field of dehazing. A lot of effective attention mechanisms have been proposed to enhance hazy images. FFA-Net~\cite{ffa-net} introduced attention mechanisms and achieved impressive results in metrics like the PSNR and SSIM. MSAFF-Net~\cite{lin2022msaff} used a channel attention module and a multiscale spatial attention module to focus on areas with features related to fog.  Chen et al.~\cite{chen2024dea} proposed the Detail-Enhanced Attention Block (DEAB), which enhances feature learning by combining Detail-Enhanced Convolution and Content-Guided Attention, thereby further improving dehazing performance. Zhang et al.~\cite{zhang2019residual} proposed a Residual Nonlocal Attention Network that takes into account the uneven distribution of information in corrupted images. They designed both local and nonlocal attention blocks to extract features for high-quality image restoration. Mou et al.~\cite{mou2021cola} introduced the COLA-Net for image restoration, which combines local and nonlocal attention mechanisms to restore areas with complex textures and highly repetitive details. However, they had high complexity and slow processing during the hazing process. These methods overlook physical characteristics. To tackle this, we propose the Haze-Aware Attention Module, thereby considering the physical model in the feature space of low-resolution images. By incorporating physical priors, we obtained effective features with fewer parameters, thus leading to 
higher PSNR and SSIM values.
\subsection{Frequency-Based Image Dehazing}
Due to the convolution theorem, Fourier analysis is widely used to address various low-level vision problems. Numerous algorithms have been researched and developed from a frequency domain perspective for low-level vision issues. Some CNN-based frameworks~\cite{selesnick2005dual,yoo2018image,yang2019wavelet} have been utilized to bridge the frequency gap between blurred and {GT} image pairs.  For instance, Chen et al.~\cite{chen2021all} proposed a hierarchical desnow network based on dual-tree complex wavelet transform to reduce snow noise in images. Yang et al.~\cite{yang2020net} developed a wavelet transform-based U-Net model to replace traditional upsampling and downsampling operations.  Zou et al.~\cite{zou2021sdwnet} employed wavelet transform to divide the input into four frequency sub-bands and processed each sub-band with separate convolutions to prevent interference between different frequency parts. Yu et al.~\cite{yu2022frequency} used deep Fourier transform to handle global frequency data and reconstruct the phase spectrum under the guidance of the amplitude spectrum, which then aids in enhancing the learning of local features within the spatial domain.
Liu et al.~\cite{liu2017efficient} achieved impressive results by removing the haze effect from the low-frequency part based on the prior that haze is typically distributed in the low-frequency spectrum of its multiscale wavelet decomposition. But, these methods all add to the complexity of wavelet or Fourier transforms, thus making the computation more costly.
We have explored a more straightforward and efficient Multiscale Frequency Enhancement Module (MFEM), which enriches and emphasizes the frequencies extracted from four size receptive fields using ultralightweight learnable parameters, and it weights the features on the channel dimension, thus achieving satisfactory results.

\section{Method}\label{sec3}

\subsection{Image Dehazing}
\begin{figure}[H]
        \includegraphics[width=0.97\linewidth]{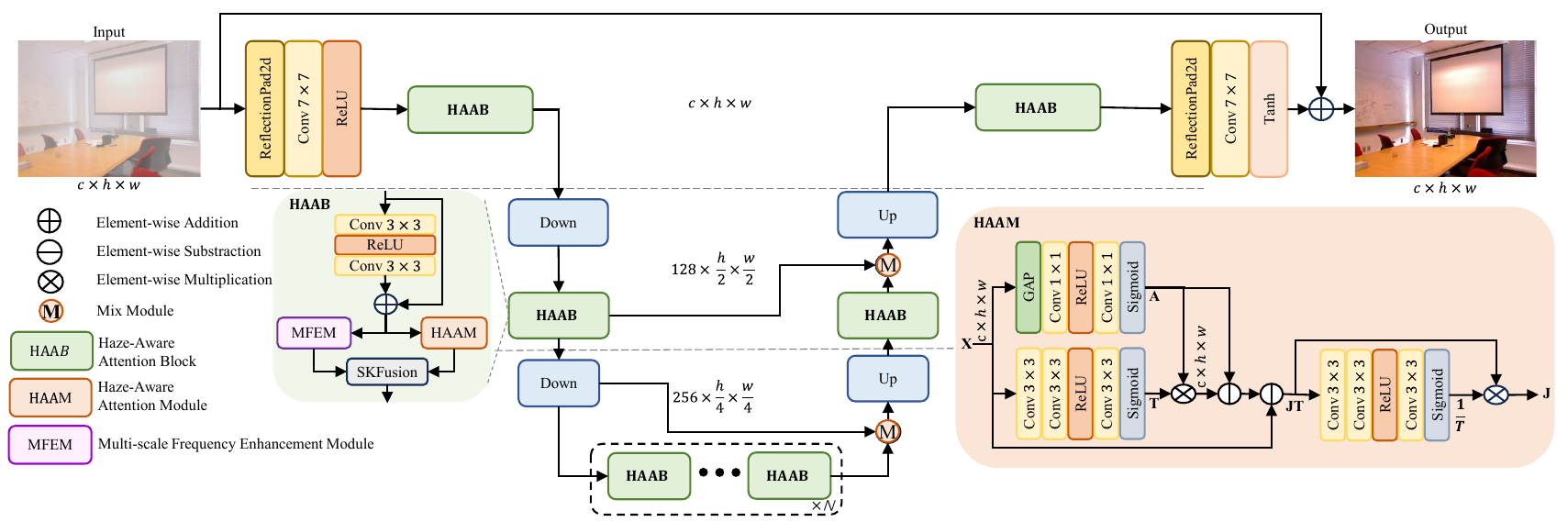}
    \caption{The 
 overview of our Haze-Aware Attention Network architecture. We give details of the structure and configurations in
Section 
 \ref{sec3}. SKFusion~\cite{li2019selective} is a feature fusion method.}
\label{network}
\end{figure}

As shown in Figure \ref{network}, our dehazing model employs a classic encoder--decoder architecture as its backbone. This framework performs a 4 $\times$ 
 downsampling operation, which greatly reduces memory usage during both training and inference, thus enhancing the model's operational efficiency. It is worth noting that our model uses three different types of activation functions. ReLU, validated in the gUnet~\cite{song2022rethinking} study for image dehazing, effectively learns complex patterns for image dehazing. Tanh, with its output range of {($-$1, 1)}, constrains the model's output to prevent extreme values, thus enhancing stability and output quality.  Additionally, the model employs a dynamic fusion module to merge features from the downsampling and upsampling layers, which is a strategy that helps to retain more information from the image and strengthens the model's ability to capture details, thus resulting in a more compact and efficient dehazing model. Within this refined feature space, we have further enhanced feature extraction and optimization through the cascade of HAABs. Each HAAB has been meticulously designed and consists of two key components: a Haze-Aware Attention Module and a Multiscale Frequency Attention Module. The HAAM, inspired by physical priors, guides the network to progressively extract clear, fog-free features, which are crucial for the dehazing effect in images. Meanwhile, the MFEM enriches the features through multiscale modulation, thus intelligently identifying and emphasizing features that contain important information and then fusing these features through channelwise weighted fusion using learnable parameters, which further improves the model's performance. With this innovative structural design, our dehazing model can effectively handle a variety of complex haze images. It not only preserves the original details of the image but also significantly improves the clarity and quality of the image.

\subsection{Haze-Aware Attention Module}
\begin{figure}[H]
\

    \includegraphics[width=0.98\linewidth]{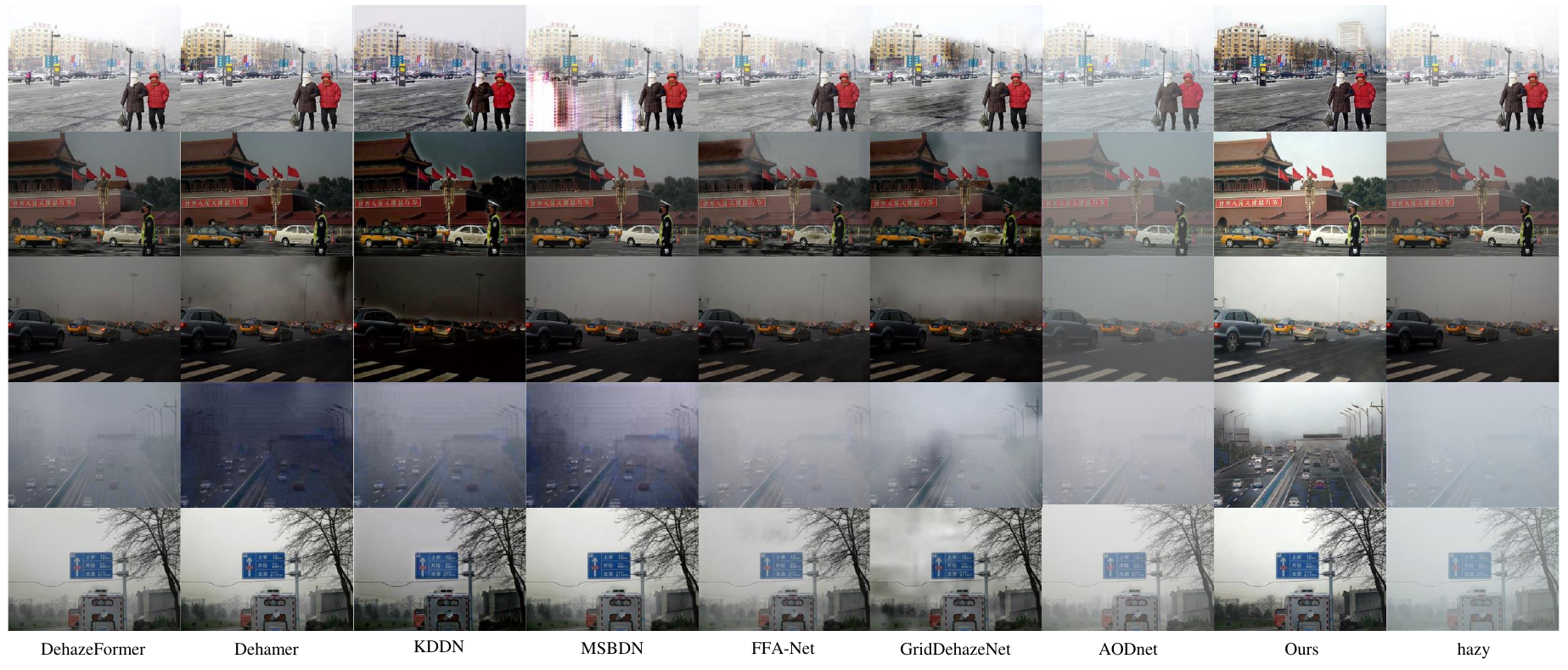}
    \caption{Visual results comparisons on real-world hazy images from the RTTS dataset~\cite{li2019benchmarking}. Zoom in for best view.}
 \label{Fig:RTTS}
\end{figure}

We introduce a new Haze-Aware Attention Module(HAAM), which cleverly applies physical models at the feature level to guide the model in feature extraction, thus pulling out a lot of potentially important features.
Empirical evidence demonstrates that this module not only provides good interpretability but also significantly improves performance metrics.
Impressively, it is likely that the introduction of the physical model makes our module more robust and adaptable for processing real images. As shown in the Figure 
 \ref{Fig:RTTS}, especially for the restoration of sky areas, it significantly outperformed other state-of-the-art methods, thus making the enhanced images more visually appealing.
Next, we will delve into the principles of the model.
Initially, leveraging the atmospheric scattering model, the generation of a hazy image can be described as follows:
\begin{equation}
    I(x) = J(x)t(x) + A(1 - t(x)),
\end{equation}
where $I$ symbolizes the hazy image, $J$ represents the GT image,  and $A$ denotes the atmospheric light, and the scattering of atmospheric light can lead to a reduction in image contrast and visibility. $T$ is the transmission map, which reflects the proportion of light that travels from a point in the scene to the camera without scattering. $X$ indicates the pixel location. The transmission map is expressed as $t = e^{-\beta d(x)}$, wherein $\beta$ signifies the atmospheric scattering coefficient, and $d$ signifies the depth of the scene. As the scene depth increases, the amount of light that reaches the camera decreases exponentially. This is because the light encounters more scattering as it passes through the atmosphere.
To streamline the process for convolution operations, we reconfigure the equation and reformulate it in a matrix representation as follows:
\begin{equation}
    JT=I-A(1-T),
\end{equation}
where $J$, $T$, $I$, and $A$ denote the matrix--vector representations of $J$, $t$, $I$, and $A$, respectively. Based on the equations presented above, we constructed the Haze-Aware Attention Module in an intuitive and effective manner. We assumed that the atmospheric light is uniform and derived $A$ by transforming the global contextual information of the entire image captured through Global Average Pooling (GAP).

\begin{equation}
A = \sigma \left( \text{Conv}^{(  \frac {N}{8},N)}\left(\text{ReLU}\left(\text{Conv}^{( N, \frac {N}{8})}\left(\text{GAP}(X)\right)\right)\right)\right),
\end{equation}

Here, $X$ represents the input feature map, $\text{GAP}(\cdot)$ signifies global average pooling, $\text{Conv}^{(N,\frac {N}{8})}(\cdot)$ refers to a convolution layer with $N$ input channels and $\frac{N}{8}$ output channels, $\sigma$ denotes the Sigmoid activation function, and $N$ is set to 64. In obtaining A, we use a process that first reduces dimensions and then increases them, which improves computational~efficiency.

Given that $\text{GAP}(\cdot)$ captures global information, thus neglecting local details and textures, we employed a $3 \times 3$ convolution layer to extract features for $T$. By introducing physical priors, this approach balances global and local features, thereby facilitating a more effective restoration of hazy images.

\begin{equation}
    T = \sigma \left( \text{Conv}^{( \frac {N}{8},N)}\left(\text{ReLU}\left(\text{Conv}^{( N, \frac {N}{8})}\left(\text{Conv}^{(N,N)}(X)\right)\right)\right)\right),
\end{equation}

Subsequently, we performed elementwise multiplication between $A$ and $(1-T)$. $JT$ was obtained via $X - A(1 - T)$. Given that division might lead to training instability, we approximated $T'$, representing $\frac{1}{T}$, using the following formula:

\begin{equation}
    T' = \sigma \left( \text{Conv}^{( \frac {N}{8},N)}\left(\text{ReLU}\left(\text{Conv}^{( N, \frac {N}{8})}\left(\text{Conv}^{(N,N)}(X)\right)\right)\right)\right),
\end{equation}

Finally, $J$ was acquired through $J = X - A(1 - T) \cdot T'$.

HAAM is an advanced attention mechanism that stands out significantly from traditional spatial and channel attention mechanisms. Its core advantage lies in its ability to integrate physical prior knowledge, thus allowing the model to learn discriminative clean features more effectively during the training process. These clean features can more accurately reflect the essence of the image, thus reducing the interference of noise and artifacts, which is crucial for high-definition image reconstruction. By integrating physical priors, HAAM not only enhances the model's understanding and processing capabilities of image content but also strengthens its generalization ability. Moreover, the design of HAAM also takes into account the computational efficiency of the model, thus reducing computational costs while improving module performance and making it a widely applicable attention~mechanism.

\subsection{Multiscale Frequency Enhancement Module}
\begin{figure}[H]
		\includegraphics[width=0.8\linewidth]{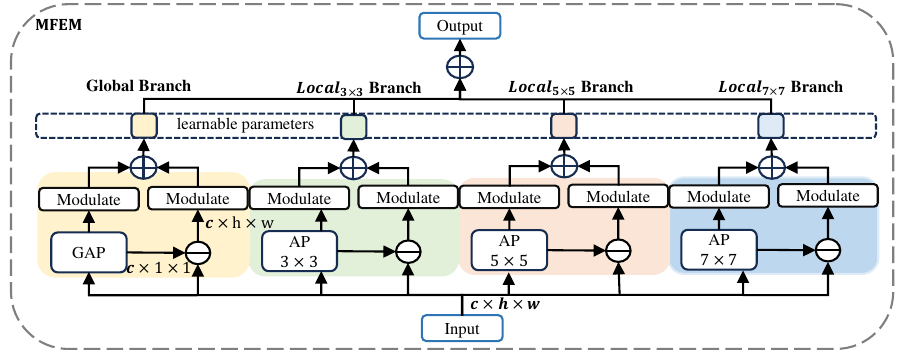}
	\caption{Multiscale Frequency Enhancement Module. GAP stands for Global Average Pooling. AP k × k means an Average Pooling operation with a kernel size of k × k. Modulation is a process that recalibrates the channels by setting attention weights as directly learnable parameters, without adding any extra layers. Learnable parameters are adjustable values that help adjust the weights at different scales.}
	\label{Fig:MFE}
	
\end{figure}

Traditional image restoration methods often focus on enhancing the frequency characteristics of images to improve their clarity and detail. These methods use transformation techniques, such as wavelet and Fourier transforms, to decompose the frequency features of an image into several different frequency bands. The purpose of this is to create isolation between signals of different frequencies, thus reducing their mutual interference, so that each band can be processed independently. Applying different convolutional kernels to these bands can further extract and enhance information within specific frequency ranges. This approach allows for the optimization of high-frequency details and low-frequency contours separately during image processing to achieve better restoration results. However, there are some limitations to this method. Firstly, it may not accurately identify and select the frequency components that carry the most important information in the image. Secondly, the need to process multiple bands separately not only increases the complexity of the algorithm but also significantly raises the computational cost. Especially during the inverse transformation, the operation must be performed separately for each band, which can be a bottleneck when computational resources are limited.
Furthermore, since the size of the degradation blur is always variable, the field of view (receptive field) is crucial in the image restoration process.  We propose an exceptionally concise and efficient multiscale frequency enhancement module that employs extremely lightweight learnable parameters to effectively decompose frequencies into distinct components, thereby highlighting parts that contain key information. As depicted in Figure \ref{Fig:MFE}, we  fully considered the impact of the receptive field in our design by utilizing convolutional kernels of sizes $3 \times 3$, $5 \times 5$, and $7 \times 7$  kernels, along with a global kernel, to capture four low-frequency components with different receptive field sizes. By subtracting these low-frequency components from the original input, we were able to generate high-frequency components and enhance the frequency sub-bands carrying significant information through network parameters. Subsequently, we applied these learnable channel weights to different frequency sub-bands. This process not only allows for individual processing of each frequency sub-band but also achieves fine-tuning of features, thus further enhancing the quality of image restoration.

\textls[+25]{The MFEM primarily consists of two main parts: the decoupler and the modulator. The decoupler acquires various frequency sub-bands using multiscale filtering. The modulator then highlights the significant frequency sub-bands with learnable parameters and processes each sub-band individually through learnable parameters on the channel~dimension.}

For any input feature map $X \in R^{C \times H \times W}$, we obtain the lowest frequency spectrum through average pooling. Then, by subtracting the low-frequency part from $X$, we obtain the high-frequency part. To fully capture spectral information from different receptive fields, we process $X$ using kernels of sizes $3 \times 3$, $5 \times 5$, and $7 \times 7$ and a global kernel. The formula is as follows:
\begin{equation}
\begin{aligned}
     X_g^l&=GAP(X),
    X_g^h=X-X_g^l, \\
    X_{3 \times 3}^l&=Conv_{3 \times 3}(X),
    X_{3 \times 3}^h=X-X_{3 \times 3}^l, \\
    X_{5 \times 5}^l&=Conv_{5 \times 5}(X),
    X_{5 \times 5}^h=X-X_{5 \times 5}^l, \\
    X_{7 \times 7}^l&=Conv_{7 \times 7}(X),
    X_{7 \times 7}^h=X-X_{7 \times 7}^l,   
\end{aligned}
\end{equation}
In
 this context, $X_g^l$ and $X_g^h$ represent the global low-frequency and high-frequency sub-bands, respectively. $ X_{3 \times 3}^l$, $X_{3 \times 3}^h$, $X_{5 \times 5}^l$, $X_{5 \times 5}^h$, $X_{5 \times 5}^h$, $X_{7 \times 7}^l$, and $ X_{7 \times 7}^h$ denote the low-frequency and high-frequency sub-bands for different receptive field sizes. To emphasize the frequency sub-bands that carry important information, we apply learnable weight parameters to the obtained frequency sub-bands for weighting.
Taking only the global receptive field as an example, the formula is as follows:
\begin{equation}
    \tilde{X_g^l}=M_g^lX_g^l,
\end{equation}
$\tilde{X_g^l}$ represents the global low-frequency sub-band after emphasizing important information. Finally, we modulate the weighted frequency sub-bands along the channel dimension using learnable parameters. The final output of the MFEM is obtained by summing these elements together. 
\begin{equation}
\begin{aligned}
       MFEM(X)=W_g^c( \tilde{X_g^l}+\tilde{X_g^h})+W_{3 \times 3}^c( \tilde{X}_{3 \times 3}^l +\tilde{X}_{3 \times 3}^h)+ \\
       W_{5 \times 5}^c (\tilde{X}_{5 \times 5}^l+\tilde{X}_{5 \times 5}^h)+W_{7 \times 7}^c( \tilde{X}_{7 \times 7}^l+ \tilde{X}_{7 \times 7}^h), 
\end{aligned}
\end{equation}
$W_k^c,k = \{1 \times 1,3 \times 3,5 \times 5,7 \times 7\}$ represents the channel attention weight maps from the filters of various scales.

Our MFEM excels at addressing uneven fog densities and irregular shapes in images, thus successfully achieving the goal of high-quality image reconstruction. This module, with its innovative multiscale processing approach, can accurately identify and handle various details and textures within the image, thus maintaining clarity and realism even under complex conditions. The core strength of MFEM lies in its fine control over different frequency components, thus allowing it to provide customized treatment for every detail in the image. By separately optimizing low-frequency and high-frequency information, the MFEM can significantly enhance the clarity of edges and textures while preserving the overall structure of the image. Moreover, the lightweight design of the module also means it has a significant advantage in computational efficiency, thereby enabling it to quickly process a large amount of image data without compromising performance.

For the loss function,
We designate the dehazed image, the clear ground truth $J_{gt}$, and the hazy image $I$ as the anchor, positive sample, and negative sample, respectively: $L_{CR}=CR( \text{HAA-Net(I)},J_{gt},I)$. Finally, we combine the loss obtained from Contrastive Regularization with the $L1$ loss function to form the final loss function:
\begin{equation}
    L_{total}=\lambda L_{CR} + L_1(HAA-Net(I), J_{gt}).
\end{equation}
{Our} experimental validation shows that when $\lambda_1=0.5$, excellent metrics were achieved.

\section{Experiments}
\label{Experiments} 
\subsection{Implementation Details}
We used the PyTorch 1.11.0 version on Four NVIDIA RTX 4090 GPUs  
to conduct all the experiments. When training, the images were randomly cropped to $ 320 \times 320$  patches. When calculating model complexity, we set the size to $128 \times 128$.  We used the Adam optimizer with a decay rate of 0.9 for $\beta_1$ and 0.999 for $\beta_2$. The starting learning rate was set at 0.00015, and we scheduled it with a cosine annealing strategy. The batch size was set to 64. Empirically, we set the penalty parameter $\lambda$ to 0.2 and $\gamma$ to 0.25, and we trained for 80~k steps. We employed Contrastive Regularization (CR)~\cite{wu2021contrastive} to better restore dehazed images. 
\subsection{Datasets and Metrics}
We used the PSNR and SSIM to evaluate the performance of our HAA-Net. We trained and tested the network on five datasets:RESIDE-Indoor
~\cite{zhong2014mining}, Haze4K~\cite{liu2021synthetic}, RTTS~\cite{li2019benchmarking}, RESIDE-Outdoor~\cite{zhong2014mining}, NH-HAZE~\cite{9150807}, and Dense-Haze~\cite{ancuti2019dense}. Specifically, the RESIDE-Indoor dataset has a total of 13,990 image pairs. We trained our model using 13,000 of those pairs and then tested the model on an additional 990 images from the RESIDE-Indoor set. We also conducted training and testing on the {RESIDE-Outdoor} dataset, which is larger and offers a more diverse set of data. This process fully demonstrates the model's generalization capabilities. The Haze4K dataset comprises 4000 
 image pairs, with 3000 used for training and the remaining 1000 for testing. Compared to RESIDE-Indoor, Haze4K  includes both indoor and outdoor scenes, thus making it more realistic. The RTTS  dataset consists of 1000 real haze images, which is ideal for assessing the generalization of our model trained on RESIDE-Indoor and Haze4K. It differs significantly from the other two datasets, thus providing a challenging and effective benchmark for evaluating the performance of HAA-Net. The NH-HAZE dataset is made up of 55 image pairs, with 50 pairs used for training and 5 pairs for testing. This setup thoroughly demonstrates our model's ability to handle fog with uneven distribution and varying densities. The Dense-Haze dataset comprises 55~pairs, including hazy images of varying sizes and densities along with their corresponding GT images. We utilized 50 pairs for training and reserved 5 pairs for testing, thereby validating the robustness of our model.

\subsection{Comparison with State-of-the-Art Methods}
\begin{figure}[H]
    
    \includegraphics[width=1\linewidth]{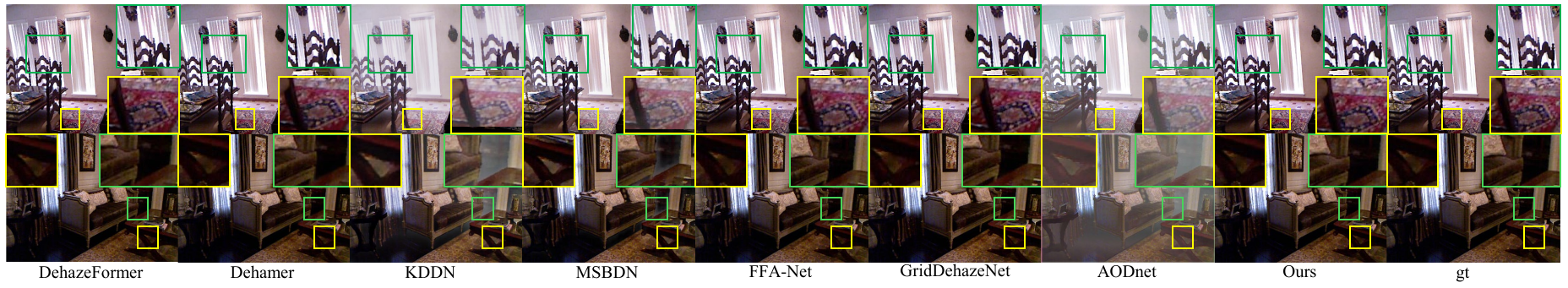}
    \caption{ Visual results comparisons on RESIDE-Indoor~\cite{li2019benchmarking} dataset.  Zoom in for best view.}
 \label{Fig:indoor}
\end{figure}

\begin{figure}[H]
    
    \includegraphics[width=1\linewidth]{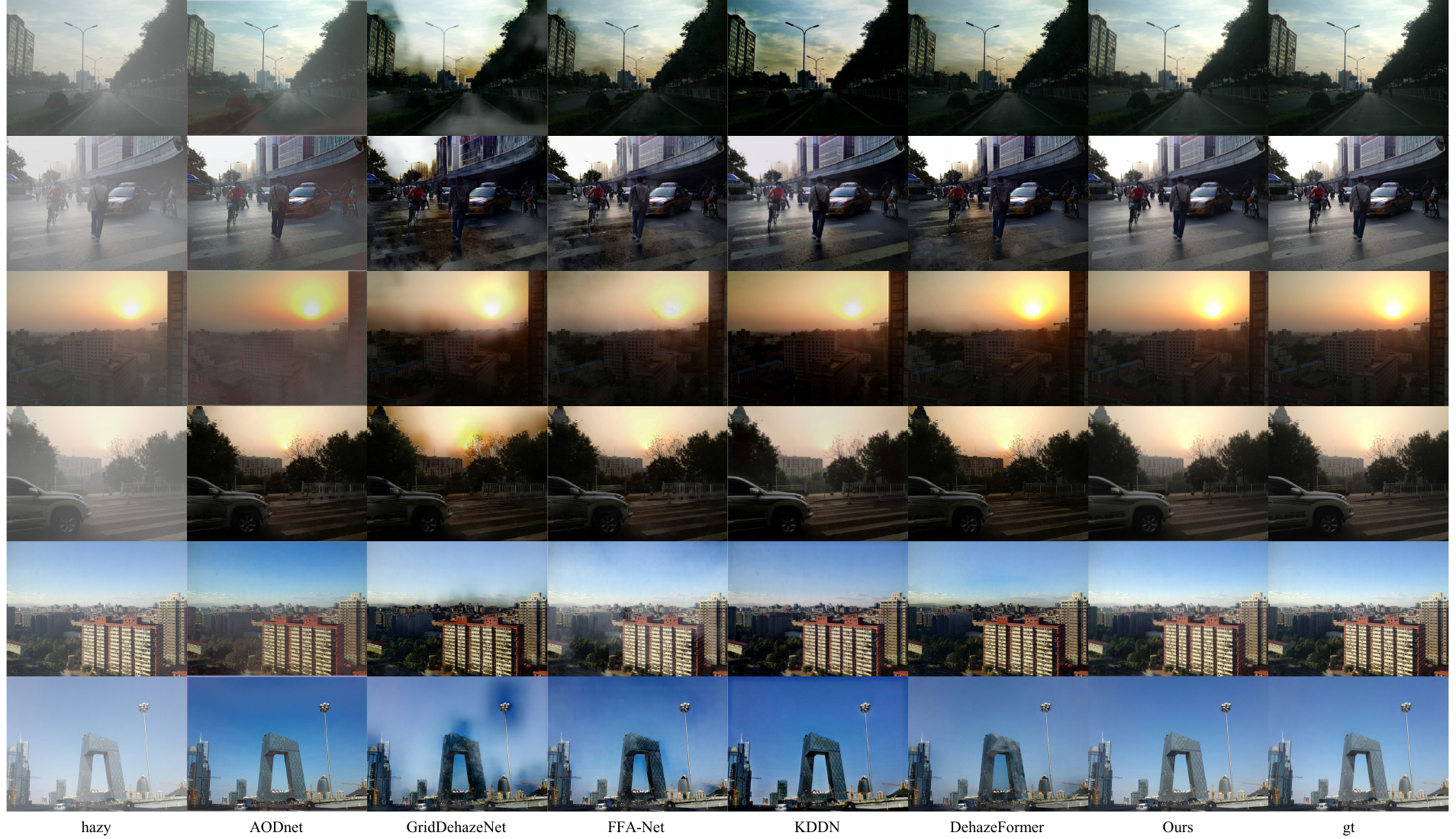}
    \caption{ {Visual} results comparisons on synthetic hazy images from the RESIDE-Outdoor dataset~\cite{li2019benchmarking}. Zoom in for best view.}

 \label{Fig:OTS}
\end{figure}

Results on Synthetic Dataset.
We compared our approach with the state-of-the-art on simulated haze images from the RESIDE-Indoor dataset, Haze4K dataset, and RESIDE-Outdoor dataset. For the RESIDE-Indoor dataset, visually, we can observe that the KDDN~\cite{hong2020distilling}, MSBDN~\cite{msbdn}, and AOD-Net~\cite{aod} suffered from texture details loss and color distortion when dealing with small patches of haze (yellow box in Figure \ref{Fig:indoor}). They also exhibited edge distortion issues (green box in Figure \ref{Fig:indoor}). While DehazeFormer-L~\cite{song2023vision}, Dehamer~\cite{guo2022image}, and FFA-Net~\cite{ffa-net} produced improved images, they sometimes overly brightened the images, thus leading to the darkening of certain details (yellow box in Figure \ref{Fig:indoor}) and showed slight edge distortion (green box in Figure \ref{Fig:indoor}). In contrast, our method excelled in preserving details, the clarity of textures, and color authenticity. For the RESIDE-Outdoor dataset, as shown in Figure 
 \ref{Fig:OTS}, you can clearly see that AOD-Net~\cite{aod} and GridDehazeNet~\cite{griddehazenet} had a lot of haze left in the images. The FFA-Net~\cite{ffa-net} and KDDN~\cite{hong2020distilling} both had some small haze leftovers, and DehazeFormer~\cite{song2023vision} made the images too dark after enhancing them. Our HAA-Net looks the closest to the clean images. When evaluating the performance on the RESIDE-Indoor and Haze4k datasets, our HAA-Net outperformed all other state-of-the-art methods. On the RESIDE-Indoor test set, as shown in Table \ref{quantitative}, the HAA-Net achieved the highest PSNR of 41.21 dB and SSIM of 0.996, thus surpassing the second-best method by a 1.16 dB PSNR improvement while also reducing the parameter count by approximately $30\%$. On the Haze4k dataset, as shown in Table \ref{quantitative}, the HAA-Net continued to demonstrate superior performance, thus achieving a PSNR of 33.93 dB and an SSIM of 0.99.
 
 \begin{table}[H]
	\caption{Quantitative 
 comparisons with SOTA  methods on the RESIDE-Indoor~\cite{li2019benchmarking}, RESIDE-Outdoor
~\cite{li2019benchmarking},
 Haze4K~\cite{liu2021synthetic}, NH-Haze~\cite{9150807}, and Dense-Haze~\cite{ancuti2019dense} datasets. Bold font indicates the optimal value for vertical comparison.} 
	\label{performance_SOTS}
	
\tablesize{\fontsize{6}{6}\selectfont}
		\begin{adjustwidth}{-\extralength}{0cm}
		\newcolumntype{C}{>{\centering\arraybackslash}X}
		\begin{tabularx}{\fulllength}{Ccccccccccccc}
			\toprule

		\multirow{2}{*}{\textbf{Method}\vspace{-6pt}} 
  & \multicolumn{2}{c}{ {\textbf{RESIDE-Indoor}}~\cite{li2019benchmarking}}  
    & \multicolumn{2}{c}{\textbf{RESIDE-Outdoor}~\cite{li2019benchmarking}} 
  & \multicolumn{2}{c}{\textbf{Haze4k}~\cite{liu2021synthetic}}   
  & \multicolumn{2}{c}{\textbf{NH-Haze}~\cite{9150807}} 

  & \multicolumn{2}{c}{\textbf{Dense-Haze}~\cite{ancuti2019dense}} 
  & \multirow{2}{*}{\textbf{{\#} Param}\vspace{-6pt}} 
  & \multirow{2}{*}{\textbf{\# MACs}\vspace{-6pt}} \\ \cmidrule{2-11}
		& \multicolumn{1}{c}{\textbf{PSNR (dB)}} & \textbf{SSIM}      & \multicolumn{1}{c}{\textbf{PSNR (dB)}} & \textbf{SSIM}
   & \multicolumn{1}{c}{\textbf{PSNR (dB)}} & \textbf{SSIM}
    & \multicolumn{1}{c}{\textbf{PSNR (dB)}} & \textbf{SSIM}
     & \multicolumn{1}{c}{\textbf{PSNR (dB)}} & \textbf{SSIM}
   \\ \midrule
		
		(ICCV'17) AOD-Net~\cite{aod}                & 19.82                     & 0.818   &20.29 &0.876  & 17.15                     & 0.830               &15.40 &0.569   &- &-  & 0.002 M 
		&-                 \\
		(ICCV'19) GridDehazeNet~\cite{griddehazenet}          & 32.16                     & 0.984 &30.86 &0.982  & -                    & -       &13.80 &0.537    &-  &-  & {0.96 M}   &-         \\
		(AAAI'20) FFA-Net~\cite{ffa-net}                & 36.39                    & 0.989  &33.57 &0.984 & 26.96                     & 0.950      &19.87 &0.692       &- &-        & {4.68 M}       &{144.17 G}           \\
		(CVPR'20) MSBDN~\cite{msbdn}                  & 33.79                     & 0.984  &- &  & 22.99                 & 0.850    &19.23 &0.706 
 &- &-    & {31.35 M}   &{20.79 G}              \\
		(CVPR'20) KDDN~\cite{hong2020distilling}           & 34.72 & 0.985  &- &-  & - & -    &17.39 &0.590  &- &-  & {5.99 M}  &-                 \\ 
		(CVPR'21) AECR-Net~\cite{wu2021contrastive} & 37.17& 0.990 &- &- & - & - &19.88 &0.717  &15.80 &0.466  & {2.61 M} &{13.05 G} \\ 
            (CVPR'22) Dehamer~\cite{guo2022image} & 36.63& 0.988 &35.18 &0.986 &  - & - &20.66 &0.684  &16.62 &0.560  &{132.45 M} &{29.57 G} 
            \\
            (ECCV'22) PMNet~\cite{ye2021perceiving} & 38.41& 0.990 &- &- &33.49 &0.980  &- &-  &- &- &{18.90 M} &-
            \\

            (TIP'23) DehazeFormer-L~\cite{song2023vision} & 40.05& \textbf{0.996} 
&- &- &32.19 &0.980  &- &-
            &-  &- &{25.44 M} &{69.93 G}
            \\
            (TIP'23) TUSR-Net~\cite{song2023tusr} & 38.67& 0.991 &- &- & - & - &- &- &18.62 &0.560
            &{5.62 M} &-
            \\
            (AAAI'24) OKNet-S~\cite{cui2024omni} &37.59 &0.994 & 35.45 & \textbf{0.992}  &- &- & 20.29 &\textbf{0.800}  &16.85 &\textbf{0.620} &{2.40 M} &{8.93 G}
            \\ (IEEE Trans. Instrum'24) MSPD-Net~\cite{yin2024multi} &39.88 &0.994  &- &- &32.97 &0.987 &- &- &- &- &-
&-

  \\ \midrule
		HAA-Net (Ours)         & \textbf{41.21}                 & \textbf{0.996} &\textbf{35.67} &\textbf{0.992}
  & \textbf{33.93}            & \textbf{0.990}  &\textbf{21.32} &0.792 
  &\textbf{18.74} &\textbf{0.620} & {18.70 M}  &{122.48 G}                \\ 
	\bottomrule
		\end{tabularx}
	\end{adjustwidth}

\label{quantitative}
\end{table}
\vspace{-9pt}

\vspace{-6pt}

\begin{figure}[H]
    
    \includegraphics[width=0.99\linewidth]{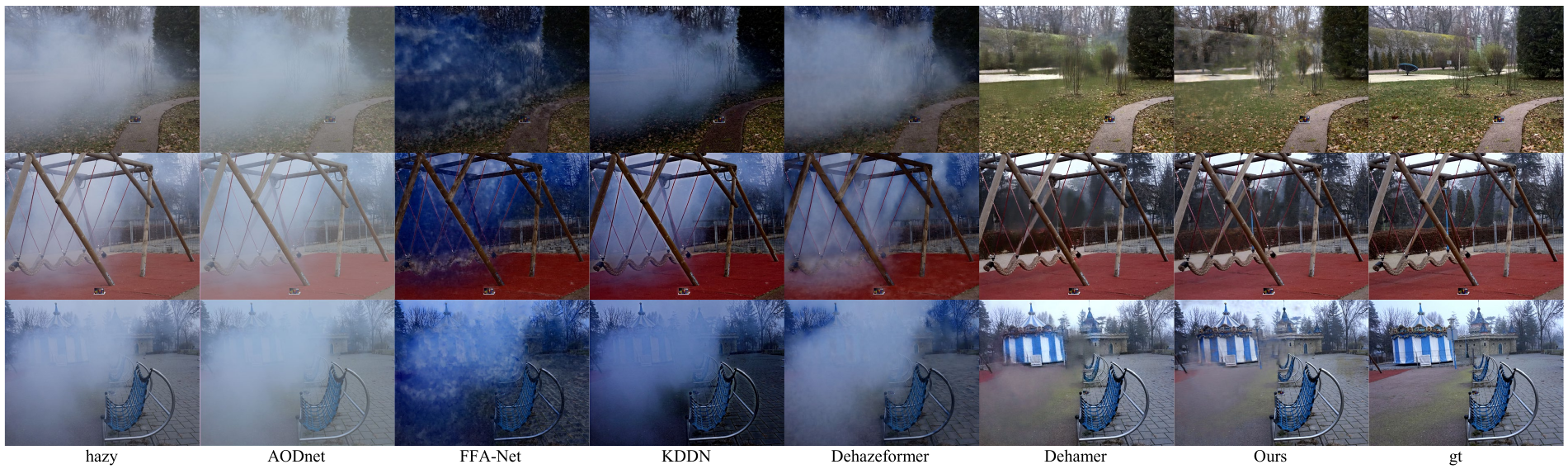}
    \caption{Visual results comparisons on real-world hazy images from the NH-HAZE dataset~\cite{9150807}. Zoom in for best view.}
 \label{Fig:NH_HAZE}
\end{figure}

Real-World Visual Comparisons.We performed real-world haze tests using samples from both the RTTS, NH-HAZE, and Dense-Haze datasets, which are more challenging than synthetic ones. The RTTS dataset includes dense and uneven haze, thereby really testing the robustness and effectiveness of the model. As shown in Figure  \ref{Fig:RTTS}, AOD-Net~\cite{aod} had a lot of large fog remnants, and both GridDehazeNet~\cite{griddehazenet} and FFA-Net~\cite{ffa-net}  had quite a bit of fog left over, with overenhancement in the sky areas. The MSBDN~\cite{msbdn}, KDDN~\cite{hong2020distilling}, and Dehamer~\cite{guo2022image} all had residual fog, and when the fog was heavy, the enhanced images turned out too dark. DehazeFormer-L~\cite{song2023vision} also had residual fog and severe texture loss. Clearly, the images restored by our HAA-Net are clear in texture and realistic in color, thus closely matching the clean images. This fully demonstrates the superior robustness and effectiveness of our method. The NH-HAZE is a {nonhomogeneous}  real image dehazing dataset. As shown in Figure \ref{Fig:NH_HAZE}, by comparing side by side, it is clear that our HAA-Net could perfectly adapt to haze of different concentrations. And our HAA-Net achieved a PSNR of 21.32 dB and an SSIM of 0.692, which is significantly better than other state-of-the-art methods. The Dense-Haze dataset includes images with various extents and densities of haze, which poses a more significant challenge for dehazing. Nonetheless, our HAA-Net method has surpassed the best-performing methods to date, thus obtaining a PSNR of 18.74 dB and an SSIM of 0.620. These achievements strongly validate that our HAA-Net is capable of effectively handling haze of different magnitudes and concentrations.

\subsection{Ablation Study} 
We performed an ablation study on our HAA-Net using the Haze4k dataset, thus gradually beefing up the key parts of the model to really show how effective each module is, as shown in Table 
 \ref{ablation_netwrok}. We started with a ``Base'' model, which is a straightforward U-Net structure with basic $3\times3$ depth convolutions. When the HAAM was added to the model, its performance improved significantly, thus achieving a PSNR of 31.76 dB and an SSIM of 0.97. What is truly impressive is that the PCNR increased by 6.3 with the addition of the HAAM, thus demonstrating the effectiveness of the HAAM module for dehazing images. As we continued to enhance the HAAM, we observed the PSNR rising even higher to 32.32~dB, along with an SSIM of 0.98. When we integrated the MFEM with the HAAM, the model truly excelled, thus achieving a PSNR of 33.46 dB and an SSIM of 0.99. Furthermore, by incorporating SKFusion technology, we elevated the PSNR to a new peak of 33.93 dB while maintaining an SSIM of 0.99. These outcomes not only validate the effectiveness of the modules we have developed but also establish HAA-Net as a standout performer in the field of image dehazing.

\begin{table}[H]
	\caption{Ablation study of our HAA-Net on the Haze4k Dataset~\cite{liu2021synthetic}.}
	\label{ablation_netwrok}
	
\tablesize{\small}
	
	\newcolumntype{C}{>{\centering\arraybackslash}X}
\begin{tabularx}{\textwidth}{cCCCC}
\toprule

		\textbf{Model} & \textbf{PSNR (dB)} & \textbf{SSIM} &    \textbf{\# Param} &\textbf{\# MACs} \\
		\midrule
            Base (U-Net)   &25.46 &0.91 & {0.85 M} &{13.35 G}\\
		Base + 
HAAM   &31.76 &0.97 & 8.6 M    &122.29 G \\
		Base + MFEM  &32.32 &0.98 & 18.2 M    &61.90 G\\
            Base + MFEM + HAAM  &33.46 &0.99 & 18.6 M &122.44 G\\
		Base + MFEM + HAAM + SKFusion (Full)  &33.93 &0.99 &18.7 M &122.48 G \\

		\bottomrule
	\end{tabularx}

\end{table}

\section{Limitations}

While our method shows excellent performance, it is not without its limitations. Specifically, due to the high complexity of our HAA-Net model and the inclusion of attention mechanisms, the number of parameters is relatively high. This could lead to increased computational costs and pose challenges in situations where computational resources are constrained. Additionally, although our network's complex design is advantageous for capturing fine-grained features in hazy images, it also results in a more complex model structure. This complexity may potentially affect the model's interpretability and could require more training data to achieve optimal performance.

Unfortunately, despite delivering remarkable results, our model, with a parameter count of 18.7 million and a computational complexity of 122.48 GMacs, still requires further optimization to be deployable on embedded devices. The deployment on embedded systems will necessitate further trade-offs between the actual performance and \mbox{operational speed}.

\section{Conclusions}
In this paper, we have introduced the HAA-Net, a novel image dehazing framework that uses the U-Net structure and includes  HAABs. This HAAB is made up of two key parts: the Haze-Aware Attention Module and the Multiscale Frequency Enhancement Module. The HAAM cleverly mixes in physical rules at the feature level, which helps the network pick up more useful underlying details during image restoration. It is likely that by including these physical models in our HAA-Net, we have managed to obtain some really impressive results when it comes to clearing up real-world hazy images. On another note, the MFEM focuses on pulling out frequency features using a multiscale field of view and highlights important information across different channels, thus making it great for dealing with fog of all sizes and densities. We put our model to the test on both made-up and real-world datasets, and the thorough evaluations really show that the HAA-Net is robust and effective for all kinds of dehazing tasks. It is clear that our method outperforms other state-of-the-art methods, thus proving its potential as a leading solution in the fields of image processing and \mbox{computer vision}.



\vspace{6pt}

\authorcontributions{Conceptualization, L.T. and E.C.; data curation, L.T.; formal analysis, Y.L.; methodology, L.T.; resources, L.C.;  software, L.T. and W.L.; supervision, Y.L. validation, Y.L., W.L. and L.C.;  writing---original draft preparation, L.T., E.C. and Y.L.  All authors have read and agreed to the published version of the manuscript.}

\funding{This
 research was funded by the Youth Science and Technology Innovation Program of Xiamen Ocean and Fisheries Development Special Funds (23ZHZB039QCB24), Xiamen Ocean and Fisheries Development Special Funds (22CZB013HJ04), the National Natural
Science Foundation of China (Grant No. 62301453).}

\institutionalreview{Not applicable.}

\informedconsent{Not applicable.}

\dataavailability{ Publicly available datasets were analyzed in this study. These data can
 be found here: \url{https://sites.google.com/view/reside-dehaze-datasets} (accessed on 22 April 2019), \url{https://pan.baidu.com/share/init?surl=41MW0YAvjFcydlroQZZizA} (accessed on 6 August 2021).}

 \conflictsofinterest{The authors declare no conflict of interest.
 } 







\begin{adjustwidth}{-\extralength}{0cm}

\reftitle{References}

\PublishersNote{}
\end{adjustwidth}
\end{document}